\documentclass[lettersize,journal]{IEEEtran}
\usepackage{amsmath,amsfonts}
\usepackage{algorithmic}
\usepackage{algorithm}
\usepackage{array}
\usepackage[caption=false,font=normalsize,labelfont=sf,textfont=sf]{subfig}
\usepackage{textcomp}
\usepackage{stfloats}
\usepackage{url}
\usepackage{verbatim}
\usepackage{graphicx}
\usepackage{cite}

\usepackage{booktabs}
\usepackage{makecell}
\usepackage{tabularx}

\usepackage{tikz}
\usepackage{pgfplots}
\pgfplotsset{compat=1.16}
\usepgfplotslibrary{groupplots}

\pgfplotsset{
    /pgf/smooth/monotonic tension/.style={
        tension={%
            if coordindex==0 then 0.3 else  % 起始点低张力
            if coordindex==1 then 0.5 else  % 第二点过渡张力
            0.7                             % 稳定区高张力
        }
    }
}

\begin{document}

\title{On-Device Training of PV Power Forecasting Models in a Smart Meter for Grid Edge Intelligence}

\author{Jian~Huang,
        Yongli~Zhu,
        Linna~Xu,~Zhe~Zheng,~Wenpeng~Cui,~Mingyang~Sun
\thanks{J. Huang, Y. Zhu and L. Xu are with Sun Yat-sen University (email: yzhu16@alum.utk.edu). Z. Zheng, W. Cui and M. Sun are with Beijing SmartChip Microelectronics Technology Co., Ltd. (email: zhengzhe@sgchip.sgcc.com.cn) (\textit{Corresponding author: Yongli Zhu})}% <-this % stops a space
%\thanks{}%
}

\markboth{Journal of \LaTeX\ Class Files,~Vol.~XX, No.~X, XXXX~202X}
{Shell \MakeLowercase{\textit{et al.}}: A Sample Article Using IEEEtran.cls for IEEE Journals}


\maketitle

% --- 摘要和关键词 (这部分会被 \maketitle 包含在单栏标题区内) ---
\begin{abstract}
In this paper, an edge-side model training study is conducted on a resource-limited smart meter. The motivation of grid-edge intelligence and the concept of on-device training are introduced. Then, the technical preparation steps for on-device training are described. A case study on the task of photovoltaic power forecasting is presented, where two representative machine learning models are investigated: a gradient boosting tree model and a recurrent neural network model. To adapt to the resource-limited situation in the smart meter, “mixed”- and “reduced”-precision training schemes are also devised. Experiment results demonstrate the feasibility of economically achieving grid-edge intelligence via the existing advanced metering infrastructures.
\end{abstract}


% --- 正文从这里开始，自动进入双栏格式 ---
\section{Introduction}
\IEEEPARstart{T}{he} penetration of edge-side (i.e, near customer side) smart devices (e.g., smart meter or concentrator) in the power grid has increased in the last ten years due to technological advancements and hardware-cost reductions. By such boosting of edge-side smart devices, the operation of the distribution power system or microgrid is expected to be \textit{edge autonomous}: some algorithms can be deployed on local, edge device(s), hence critical decisions can be made and send promptly with less reliance on the the remote  (cloud) control center -- also called ``grid edge intelligence”. 

Regarding the applications of edge intelligence, the ``on-device inference” scheme has prevailed in the fields of IoT (internet-of-things), auto-driving, and power grid. For instance, in \cite{ref1}, the author deploys a trained object-detection model on several edge computing devices with milliwatt power consumption. In \cite{ref2}, the gradient-boosting tree, linear regression, and support vector machine models are respectively deployed in a smart meter for photovoltaic (PV) voltage control. 

On the other hand, driven by the need for communication cost reduction, agile model updating, and privacy preservation, the idea of ``on-device training”, i.e., training machine learning (ML) models \textit{directly} in edge devices, has emerged in recent IoT applications. In \cite{b1}, an IoT device with only 256KB of memory is used for on-device training. In \cite{b2}, the author uses FP8 (8-bit float numbers) for on-device training of federated-learning models. In \cite{b3}, the author presents a so-called Tiny-Transfer-Learning method that significantly reduces the memory footprint of on-device training by freezing the weights of the pre-trained model and updating only the bias.

In view of the above technology trends and the emerging need of grid edge intelligence, this paper investigates the feasibility of the ``on-device training” idea in a real smart meter, where two representative ML models, i.e., XGBoost (Extreme Gradient Boosting) and LSTM (Long Short-Term Memory), are trained for a PV power forecasting task based on real microgrid data. Besides, \textit{mixed}- and \textit{reduced}-precision training schemes are leveraged in training the LSTM models. 
% Section \ref{sec2} describes the basic workflow, the task specification, and the hardware description. Section III presents the detailed preparation steps for implementing the on-device training idea in the smart meter. Section IV presents case study results on a PV power forecasting dataset from a real microgrid, where the \textit{reduced-precision training} technique is adopted for deep learning models. 

\begin{figure}
    \centering
    \includegraphics[width=1\linewidth]
    {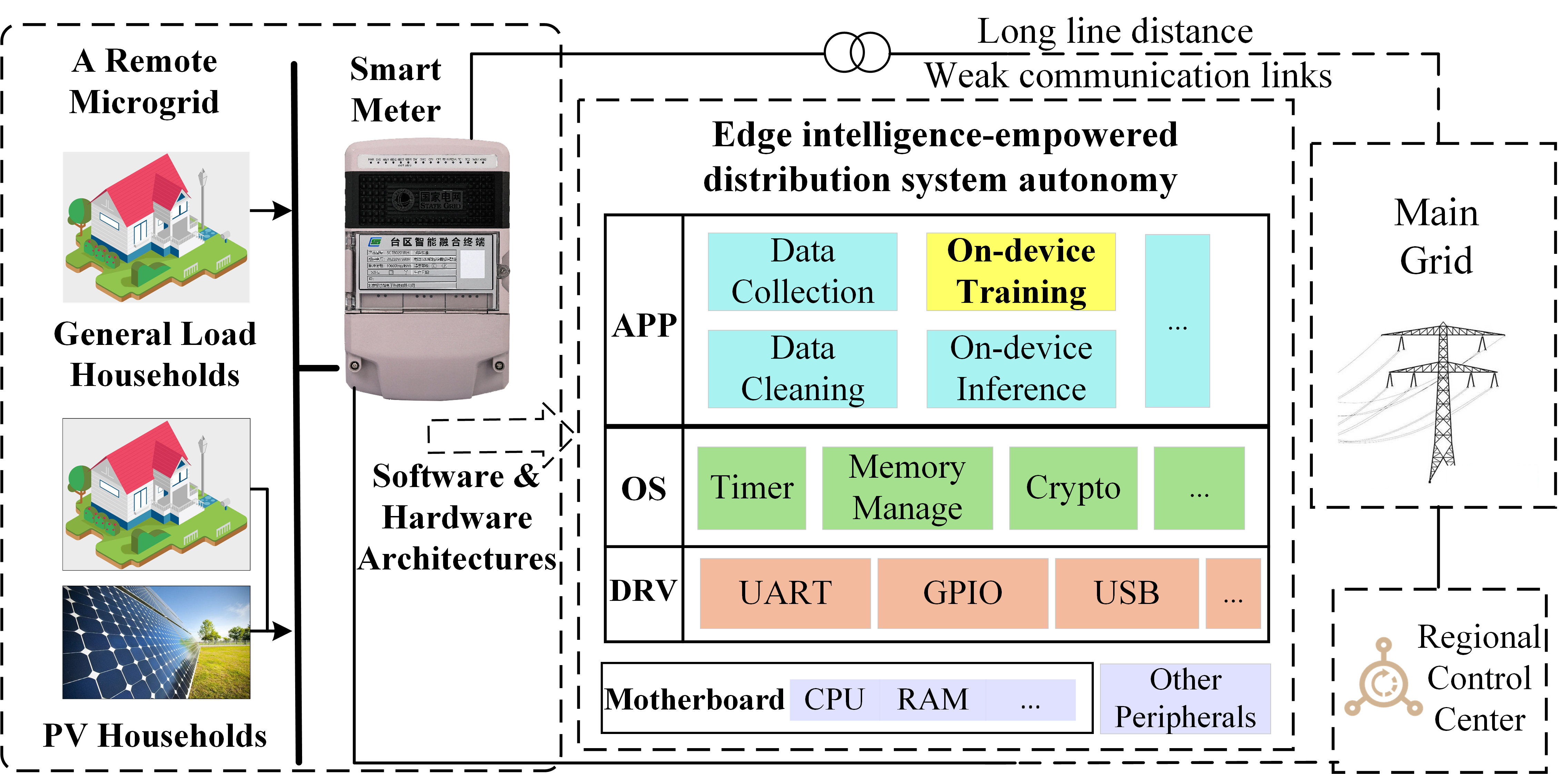}
    \caption{The basic idea of ``on-device training” for PV power forecasting}
    \label{fig1}
    \vspace{-14pt}
\end{figure}

% \vspace{-2pt}

\section{Descriptions of the Meter and the Task}\label{sec2}

\subsection{Software and Hardware Specifications of the Smart Meter}
Fig. \ref{fig1} displays the meter device used in this work, where model training is performed. The meter is based on an ARM board with a quad-core Cortex-A53 CPU, 984MB memory, and 977MB disk space (where 478MB is available for our development after excluding the system reserved space). The operating system (OS) is 64-bit ARM-Linux 4.9. %The GCC version is 7.5.% 
It is worth mentioning that this meter is a mass-production type that is \textit{not specially} designed for on-device training.

\vspace{-3pt}

\subsection{PV Power Forecasting Task in a Remote Microgrid}
Fig. \ref{fig1} describes a use-case of on-device training for grid-edge intelligence in the remote microgrid scenario: unstable communication with the regional control center due to poor infrastructure; no energy storage within the microgrid due to affordability; weak support from the main grid due to long electrical distance. In such a scenario, on-device model training of the PV power forecaster can help the local owner of the PV inverter or microgrid make timely control decisions.

% This paper uses the PV power forecasting task as a concrete demonstration of on-device training and grid-edge intelligence. 
Due to the unavailability of the weather data (e.g., irradiation), only the historical power measurements are used here as the input features. Hence. the forecasting task can be formulated by Eq. (\ref{eq1}), where \(k\) is the feature length, \(h\) is the forecasting horizon, and $f$ represents a specific ML model.

\vspace{-9pt}
\begin{equation}
\label{eq1}
P_{t+h}=f(P_{t},P_{t-1},P_{t-2},...,P_{t-k})
\end{equation}

% \begin{figure}
%     \centering
%     \includegraphics[width=0.75\linewidth]{Meter5.jpg}
%     \caption{The ARM board of the smart meter used in the experiment}
%     \label{Main Control Unit of Photovoltaic Power Grid}
% \end{figure}

\section{Deploying Model-Training Components on the Smart Meter}
In addition to the previously mentioned resource-limited conditions of the smart meter, the following challenges need to be tackled for implementing the on-device training idea:
\begin{itemize}
    \item Existing ML developing tools (e.g., sklearn, PyTorch, and TensorFlow) lack official support for on-device-training libraries (e.g., \textit{back-propagation} or \textit{auto-differentiation}).
    \item Existing meter OS lacks \textit{off-the-shelf} runtime components for model-training frameworks; thus, manual cross-compilation or customized development is needed.
    % \item Training a neural network model often requires large on-chip storage (ROM) and memory (RAM).
\end{itemize}

\begin{figure}
    \centering
    \includegraphics[width=1\linewidth]
    {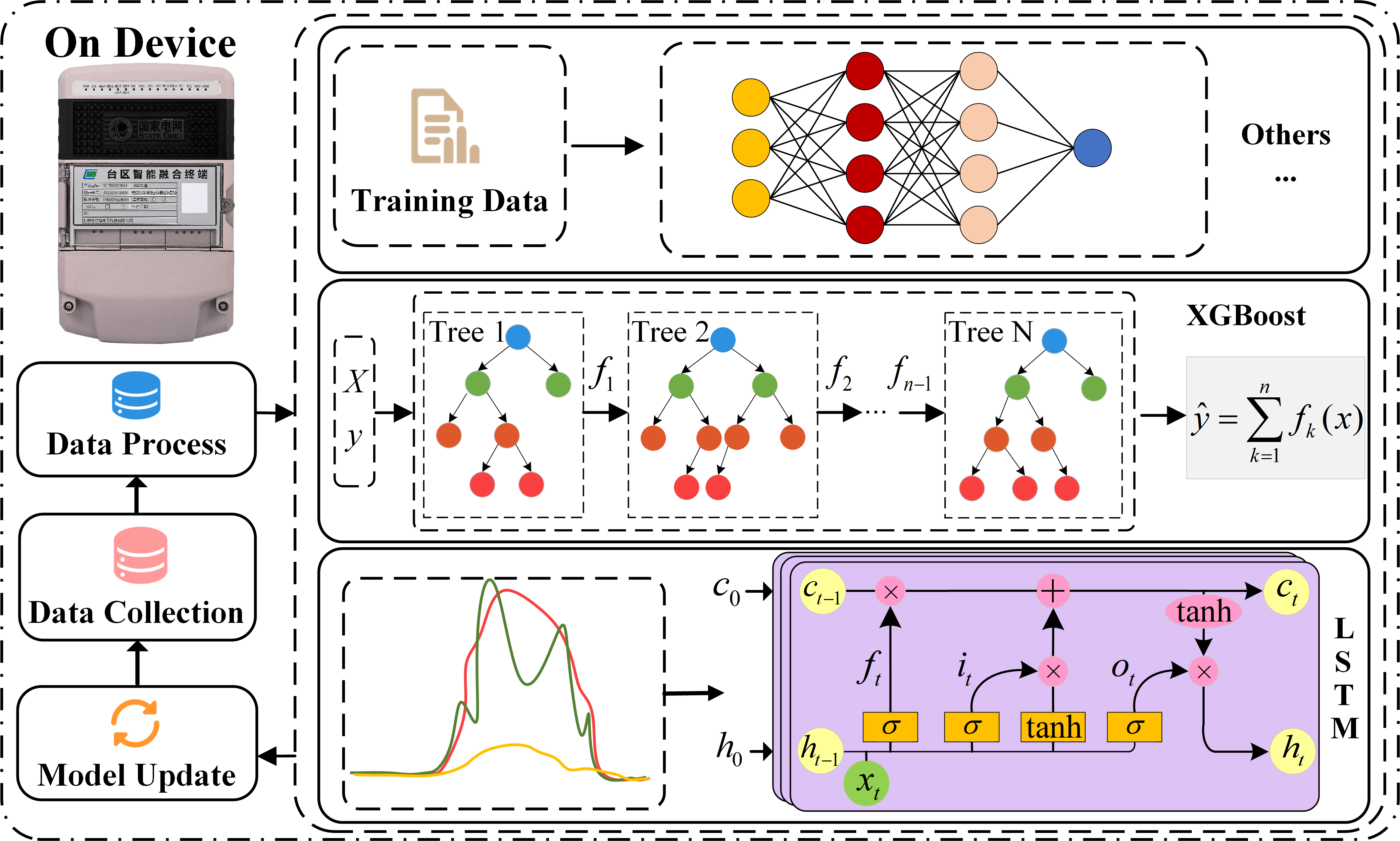}
    \caption{The workflow for on-device training of two representative ML models}
    \label{fig2}
    % \vspace{-12pt}
\end{figure}

To inspect the performance of the on-device training approach in our resource-limited smart meter, two representative ML models are considered: XGBoost (a \textit{widely} used non-deep learning model in time-series competitions) and LSTM (a \textit{classic} deep learning model for time-series forecasting), and the workflow is shown in Fig. \ref{fig2}. The subsections below detail the preparation steps for training the two models in the meter.

% \vspace{-5pt}

\subsection{Cross-compilation of Python}\label{AA}
% We use Python to implement XGBoost on-device training, so naturally Python and its XGBoost-related libraries (such as Numpy, pandas, and Scipy) needed to be cross-complied (one technique can compile executable files for target devices suitable for another architecture on the host of one architecture) and installed on the ARM board and gcc-linaro (version 7.5) is chosen as the cross-complication tool. In order to cross-compile these libraries, we first need to obtain the source code, then use the cross-compile toolchain for cross-complication, and finally deploy the cross-complication executable file to the ARM board, these cross-complications will encounter some problems such as compatibility of dependency libraries, configuration errors in complication options, etc., which will not be described here.

XGBoost provides Python and C++ interfaces, and we found that the time and accuracy of training XGBoost using the Python interface are satisfactory for the PV forecasting task in certain pre-experiments. Therefore, the Python interface is adopted to train XGBoost. To that end, the cross-compilation and deployment of the Python language and necessary dependencies are needed. Several key compilation steps are shown in Table \ref{T1}, where \textit{zlib}, \textit{openssl}, \textit{libffi}, and \textit{bz2} must be cross-compiled before cross-compiling Python from source.

\begin{table}[htbp]
% \vspace{-6pt}
\centering
\caption{Key Compilation Steps Needed for Cross-compiling Python}
\label{T1}
\renewcommand{\arraystretch}{1.2}
\setlength{\tabcolsep}{8pt}
\scriptsize  % 整体缩小字号
\begin{tabular}{@{}>{\ttfamily}l c >{\ttfamily}l@{}}
\toprule
\multicolumn{1}{c}{\normalfont\textbf{Package}} & 
\multicolumn{1}{c}{\textbf{Version}} & 
\multicolumn{1}{c}{\normalfont\textbf{Compilation Commands}} \\
\midrule
\multicolumn{1}{c}{zlib} & 1.3.1 & ./configure \\
\addlinespace[2pt]
\multicolumn{1}{c}{openssl} & 1.1.1h & \begin{tabular}[t]{@{}l@{}}
./Configure linux-aarch64 \\
no-asm shared no-async
\end{tabular} \\
\addlinespace[2pt]
\multicolumn{1}{c}{libffi} & 3.3 & ./configure --host=aarch64-linux-gnu \\
\addlinespace[2pt]
\multicolumn{1}{c}{bz2} & 1.0.8 & \begin{tabular}[t]{@{}l@{}}
CC=aarch64-linux-gnu-gcc \\
AR=aarch64-linux-gnu-ar \\
PREFIX=/path/to/bz2/ \\
CFLAGS="Original Content plus fPIC”
\end{tabular} \\
\addlinespace[2pt]
% Python & 3.10 & \begin{tabular}[t]{@{}l@{}}
% ./configure --host=aarch64-linux-gnu \\
% --build=x86\_64\-linux\-gnu \\
% --enable-optimizations \\
% CPPFLAGS="-I/path/to/fourModules/include/"\\
% LDFLAGS="-L//path/to/fourModules/lib/"
% \end{tabular} \\
\bottomrule
\end{tabular}
\vspace{-10pt}
\end{table}

% \vspace{-5pt}
\subsection{Components for On-device Training of the XGBoost Model}

After compiling and installing Python, the runtime library of XGBoost and other dependency packages can be cross-compiled in the meter. Their versions are shown in Table \ref{T2}.

\begin{table}[htbp]
% \vspace{-6pt}
\centering
\footnotesize
\caption{Major Packages Needed for the XGBoost-Python Interface}
\label{T2}
\begin{tabular}{ccccccc}
\toprule
\textbf{Package} & pip  & wheel & numpy & scipy & pandas & XGBoost \\
\midrule
\textbf{Version} & 24.3.1 & 0.45.1 & 1.26.3 & 1.10.0 & 1.5.0 & 2.1.1 \\
\bottomrule
\end{tabular}
\vspace{-6pt}
\end{table}

% In traditional machine learning, model training is done on machines with powerful computing power, and machine learning models are rarely trained directly on edge devices. However, model training on desktops is already a mature topic, so this article focuses on direct model training on edge devices with weak computing power. This section illustrates how to train the XGBoost directly on the ARM board.

% XGBoost is an effective gradient-enhancing decision tree algorithm. which integrates multiple weak learners into a strong learner through certain methods. Its core idea can be expressed by the following formula.
% \begin{equation}
% \hat{y}_i=\sum_{k=1}^K f_k\left(x_i\right)
% \end{equation}
% In Eq.(2), KK represents the number of weak learners, fkf_k represents the k-th weak learner. The i-th sample data xix_i is converted by Eq.(2) to obtain the predicted result yiy_i. For more details about XGBoost, please refer to [?]. 

\subsection{Components for On-device Training of the LSTM Model}
So far, no freely available, ready-to-go libraries exist for on-device training of an LSTM model. Moreover, there is no extra space to install PyTorch/TensorFlow in our meter (even if the storage space is enough, the cross-compilation of PyTorch/TensorFlow \textit{per se} can be laborious). Thus, we choose to implement LSTM and its training framework from scratch using C++ and \textit{Eigen} \cite{b4}, a lightweight template library for speeding up matrix computation. Unlike other libraries that must first be cross-compiled towards the target platform, \textit{Eigen} can be used by simply including its header files. Besides, our program adopts a modular design, where the LSTM cell, LSTM layer, and LSTM model are separately encapsulated, making it easier for future upgrades. Table \ref{T3} lists the hyperparameters for on-device training of the LSTM.
%Eigen这个库是跟程序一起交叉编译的(Eigen这个库跟上面那些库不一样，这个Eigen库没有lib)，我并没有交叉编译Eigen本身，电表上也没有Eigen的动态库这些文件，开发板本身也没有部署c++的工具链

% \begin{table}[htbp]
% \centering
% \footnotesize  % 比 \scriptsize 略大，可根据需要改为 \small 或 \scriptsize
% \caption{The Structure and Training Hyperparameters for LSTM}
% \label{tab:comparison}
% \begin{tabular}{%
%   >{\centering\arraybackslash}p{0.45\columnwidth}
%   >{\centering\arraybackslash}p{0.45\columnwidth}}
% \toprule
% \textbf{Parameter} & \textbf{Value}\\
% \midrule
% no. of LSTM cells & 24 or 96  \\
% no. of hidden units & 32  \\
% output dim. & 1  \\
% training epochs & 50  \\
% batch size & 16  \\
% learning rate & 0.001  \\
% \bottomrule
% \end{tabular}
% \end{table}

\begin{table}[htbp]
% \vspace{-6pt}
\centering
\footnotesize
\caption{The Structural and Training Hyperparameters for LSTM}
\label{T3}
\resizebox{\columnwidth}{!}{%
\begin{tabular}{ccccccc}
\toprule
\textbf{Param.} & \makecell{num. of\\ cells}  & \makecell{hidden\\ units} & \makecell{output\\ dim.} & \makecell{training\\ epochs} & \makecell{batch\\ size} & \makecell{learning\\ rate}\\
\midrule
\textbf{Value} & 24 or 96 & 32 & 1 & 50 & 16 & 0.001\\
\bottomrule
\end{tabular}
}
% \vspace{-6pt}
\end{table}

\vspace{-4pt}
\section{Case Study}
The dataset used in this study is sourced from a village microgrid. The training set contains the historical generation of six PV households. The meter collected data from each household every 15 minutes (96 samples per day) from January 2024 to May 2024. The goal is to forecast the power output of each PV in 24 hours later (i.e., $h=96$ in Eq. (\ref{eq1})).

The following MAPE (Mean Absolute Percentage Error), as defined in Eq. (\ref{eqn2}), is used as the performance metric. $y_i$ is the true value, and $\hat{y}_i$ is the prediction. \(n\) is the total number of testing samples. $Cap$ is the capacity of a specific PV generator.

\vspace{-4pt}
\begin{equation}
\label{eqn2}
\text { MAPE }=\left(1-\sqrt{\frac{1}{n} \sum_{i=1}^n\left(\frac{y_i-\widehat{y_i}}{Cap}\right)^2}\right) \times 100 \%
\end{equation}

\subsection{Results of On-Device Training of the XGBoost Model}

The XGBoost models (with $k$ = 96 in Eq. (\ref{eq1})) are then trained on a desktop PC (5.0GHz CPU, 32GB RAM, and 4060ti GPU) and the smart meter, respectively. Comparison results regarding the training time and MAPE are listed in Table \ref{T4}. In the table, each row represents a PV household, and the total data length of each PV household is given. As observed, the testing accuracy of on-device training is similar to that of on-PC training in terms of the MAPE metrics. 

The time costs of on-device training are higher, which is unsurprising since the PC's CPU/GPU is much more powerful than the meter. However, such training time costs (around 200 sec) can still satisfy the typical need for model updating in this PV forecasting task (e.g., retraining the model hourly).
% (indeed, the on-device inference takes only milliseconds).

\begingroup
% 仅在此组内修改行距
\renewcommand{\arraystretch}{0.6}
% 仅在此组内修改 \toprule 
\setlength{\aboverulesep}{1pt}
\setlength{\belowrulesep}{1pt}
\newcommand{\headercell}[1]{{\fontsize{5pt}{6.5pt}\selectfont\makecell{#1}}}

\newcommand{\customtiny}{\fontsize{5pt}{0pt}\selectfont}

\begin{table}[htbp]
\vspace{-6pt}
\centering
\caption{Results Comparison for Training the XGBoost Model}
\label{T4}
\resizebox{\columnwidth}{!}{%
\begin{tabular}{c c| c c |c c}
\toprule
\headercell{\textbf{PV} \\ \textbf{ID}} & 
\headercell{\textbf{Data} \\ \textbf{Length}} & \headercell{\textbf{On-PC} \\ \textbf{Time (sec)}} & \headercell{\textbf{MAPE}} & \headercell{\textbf{On-Device} \\ \textbf{Time (sec)}} & \headercell{\textbf{MAPE}}\\
\midrule
{\customtiny1} & {\customtiny
12192} & {\customtiny0.72} & {\customtiny9.34\%} & {\customtiny203.57} & {\customtiny9.36\%}\\
{\customtiny2} & {\customtiny12192} & {\customtiny0.45} & {\customtiny8.22\%} & {\customtiny213.49} & {\customtiny8.05\%}\\
{\customtiny3} & {\customtiny12096} & {\customtiny0.42} & {\customtiny9.03\%} & {\customtiny204.17} & {\customtiny9.06\%}\\
{\customtiny4} & {\customtiny12096} & {\customtiny0.76} & {\customtiny9.15\%} & {\customtiny268.32} & {\customtiny9.21\%}\\
{\customtiny5} & {\customtiny11904} & {\customtiny0.95} & {\customtiny9.27\%} & {\customtiny215.30} & {\customtiny9.26\%}\\
{\customtiny6} & {\customtiny12000} & {\customtiny0.61} & {\customtiny9.48\%} & {\customtiny230.31} & {\customtiny9.47\%}\\

\bottomrule
\end{tabular}
}
\vspace{-6pt}
\end{table}
\endgroup

% \begin{table}[htbp]
% \centering
% \caption{Comparison of Different Household Data by XGBoost On-Device Training}
% \label{tab:comparison}
% \resizebox{\columnwidth}{!}{%
% \begin{tabular}{c c c }
% \toprule
% \makecell{\textbf{Total Data} \\ \textbf{Length}} & \makecell{\textbf{On-Device Training Time (sec)}} & \textbf{MAPE} \\
% \midrule
% 12192 & 203.57 & 9.36\% \\
% 12192 & 213.49 & 8.05\% \\
% 12096 & 204.17 & 9.06\% \\
% 12096 & 268.32 & 9.21\% \\
% 11904 & 215.30 & 9.26\% \\
% 12000 & 230.31 & 9.47\% \\
% \bottomrule
% \end{tabular}
% }
% \end{table}

\subsection{Results of On-Device Training of the LSTM Model}

Here, the 2nd PV household's data (in Table \ref{T4}) is picked to compare the LSTM's training performances on the PC and on the meter. The results are listed in Table \ref{T5}, where the ``Feature Length” column represents the value of \(k\) in Eq. (\ref{eq1}).

\begin{table}[htbp]
\vspace{-6pt}
\centering
\caption{Results Comparison for Training the LSTM Model (PV ID = 2)}
\label{T5}
\resizebox{\columnwidth}{!}
{%
\begin{tabular}{cc|cc|cc}
\toprule
\makecell{\textbf{Data} \\ \textbf{Length}} & \makecell{\textbf{Feature} \\ \textbf{Length}} & \makecell{\textbf{On-PC} \\ \textbf{Time (sec)}} & \textbf{MAPE} & \makecell{\textbf{On-Device} \\ \textbf{Time (sec)}} & \textbf{MAPE}\\
\midrule
% \customtiny{
2976 & 24 & 6 & 5.67\% & 421 & 5.69\%\\
2976 & 96 & 25 & 6.02\% & 1615 & 5.97\%\\
12192 & 24 & 26 & 9.39\% & 1774 & 9.38\%\\
12192 & 96 & 136 & 9.40\% & 7043 & 9.38\%\\
% }
\bottomrule
\end{tabular}
}
\vspace{-6pt}
\end{table}

\subsection{Mixed and Reduced-Precision Training of the LSTM Model}
In this section, two reduced precision schemes are devised to speed up the on-device training process. More specifically, (part of) the input data and defined variables are converted into a lower-precision type (float32) from the default higher-precision type (double). Two experiments are conducted: 

\begin{itemize}
\item{Convert 100\% variables to float32 (marked as ``Float”)}
\item{Convert 50\% variables to float32 (marked as ``Mixed”)}
\end{itemize}

\begin{table}[htbp]
\vspace{-6pt}
\centering
\caption{Mixed and Reduced-Precision On-Device Training of the LSTM Model (Different Conversion Percentages) (PV ID = 2)}
\label{T6}
\resizebox{\columnwidth}{!}{%
\begin{tabular}{c c c c c c}
\toprule
& & \multicolumn{2}{c}{``Float”}  & \multicolumn{2}{c}{``Mixed”} \\
\cmidrule(lr){3-4}                  
\cmidrule(lr){5-6}
\makecell{\textbf{Data} \\ \textbf{Length}} & \makecell{\textbf{Feature} \\ \textbf{Length}} & \textbf{Time (sec)} & \textbf{MAPE} & \textbf{Time (sec)} & \textbf{MAPE}\\
\midrule
2976 & 24 & 184 & 5.72\% & 265 & 5.69\%\\
2976 & 96 & 706 & 5.90\% & 1010 & 5.97\%\\
12192 & 24 & 774 & 9.06\% & 1112 & 9.38\%\\
12192 & 96 & 3074 & 8.77\% & 4400 & 9.38\%\\
\bottomrule
\end{tabular}
}
% \vspace{-10pt}
\end{table}

\begin{figure}
    \centering
    \includegraphics[width=0.94\linewidth]
    {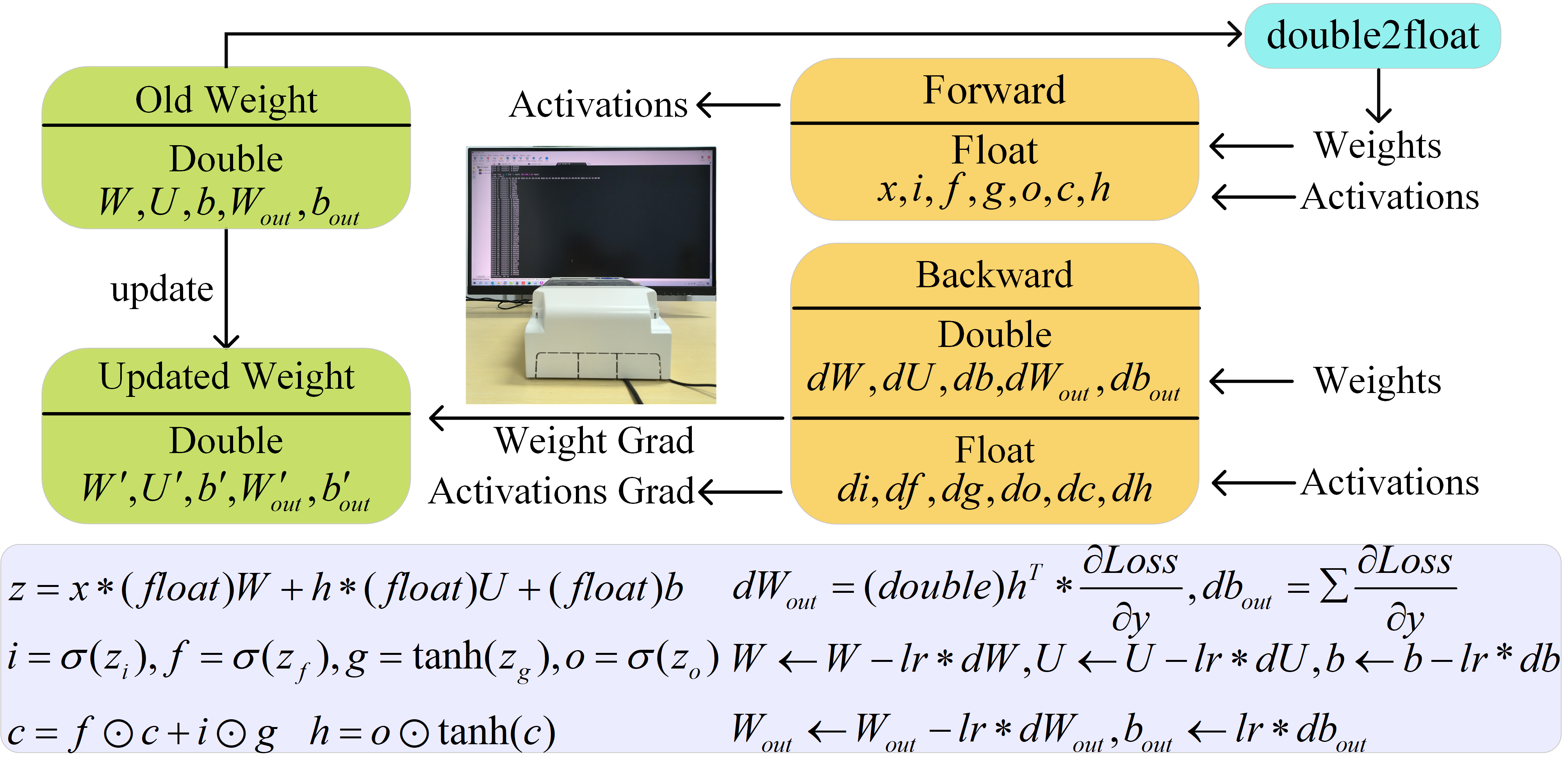}
    \caption{The workflow of the proposed mixed-precision training scheme}
    \label{fig3}
    % \vspace{-6pt}
\end{figure}

The workflow is illustrated in Fig. \ref{fig3}. Experiment results for the 2nd PV household are shown in Table \ref{T6}. It is observed that the ``Float” scheme consumes the least wall-clock time with similar accuracy as the default double precision scheme.

\begin{figure} 
% \vspace{-10pt}
\begin{center}
\begin{tikzpicture}[scale=0.71, transform shape]
    \begin{groupplot}[
      group style={
        group size=2 by 2,
        horizontal sep=0.2cm,  % 进一步减小水平间距
        vertical sep=0.1cm,    % 进一步减小垂直间距
        xticklabels at=edge bottom,
        yticklabels at=edge left, 
        x descriptions at=edge bottom,
        y descriptions at=edge left
      },
      width=0.3\textwidth,    % 减小单个子图宽度
      height=0.1\textwidth,   % 减小单个子图高度
      scale only axis,
      legend to name=commonlegend,
      legend style={
        legend columns=3,
        cells={anchor=center},
        /tikz/every even column/.append style={column sep=0.2cm},
        font=\footnotesize     % 缩小图例字体
      },
      grid=both,
      grid style={line width=.1pt, draw=gray!30},
      major grid style={line width=.2pt,draw=gray!50},
      axis background/.style={fill=white},
      every axis/.append style={
        axis line style={draw=black},
        tick label style={font=\footnotesize},  % 缩小刻度标签
        label style={font=\footnotesize}        % 缩小轴标签
      },
      xlabel style={yshift=-0.1cm},  % 调整x标签位置
      ylabel style={xshift=-0.1cm}   % 调整y标签位置
    ]
  
     % Plot 1 (左上) 
  \nextgroupplot
  \node[
    fill=gray!20,          % 浅灰色填充
    draw=none,              % 无边框
    align=left,             % 左对齐
    font=\small,            % 缩小字体
    anchor=north east,      % 锚点在右上角
    xshift=-3mm,            % 向左微调位置
    yshift=-1mm             % 向下微调位置
  ] at (rel axis cs:1,1) { % 定位在子图坐标系右上角
    Data Length: 2976\\     % 第一行
    Feature Length: 24       % 第二行
  };
  \addplot[red, mark=none ,smooth, tension=0.6,line width=1pt] coordinates {
    (1, 3.10856) (2, 2.3508) (3, 1.7677) (4, 1.36502) (5, 1.11708) (6, 0.975115) (7, 0.894979) (8, 0.84818) (9, 0.819167) (10, 0.799961) (11, 0.786443) (12, 0.776395) (13, 0.768556) (14, 0.762179) (15, 0.756802) (16, 0.752133) (17, 0.747984) (18, 0.744229) (19, 0.740783) (20, 0.737586) (21, 0.734595) (22, 0.731778) (23, 0.729108) (24, 0.726567) (25, 0.724136) (26, 0.721804) (27, 0.719558) (28, 0.717391) (29, 0.715297) (30, 0.713273) (31, 0.71132) (32, 0.709443) (33, 0.707645) (34, 0.705931) (35, 0.704305) (36, 0.702766) (37, 0.701307) (38, 0.699922) (39, 0.698602) (40, 0.697339) (41, 0.696127) (42, 0.69496) (43, 0.693835) (44, 0.692748) (45, 0.691695) (46, 0.690675) (47, 0.689685) (48, 0.688723) (49, 0.687789) (50, 0.68688)
  };
  \addlegendentry{test1}
  \addplot[green, mark=none,smooth, tension=0.6,line width=1.3pt,dotted] coordinates {
    (1, 3.28573) (2, 2.76669) (3, 2.3264) (4, 1.96105) (5, 1.66857) (6, 1.43993) (7, 1.26675) (8, 1.1392) (9, 1.04572) (10, 0.976786) (11, 0.926037) (12, 0.888363) (13, 0.859974) (14, 0.838305) (15, 0.821488) (16, 0.808124) (17, 0.797347) (18, 0.78844) (19, 0.780934) (20, 0.774452) (21, 0.768727) (22, 0.763578) (23, 0.758868) (24, 0.754497) (25, 0.750396) (26, 0.746509) (27, 0.742825) (28, 0.739331) (29, 0.736049) (30, 0.732998) (31, 0.730202) (32, 0.727674) (33, 0.725399) (34, 0.723337) (35, 0.721462) (36, 0.719754) (37, 0.718182) (38, 0.716727) (39, 0.715365) (40, 0.714086) (41, 0.712874) (42, 0.711724) (43, 0.710631) (44, 0.709581) (45, 0.708563) (46, 0.707576) (47, 0.706617) (48, 0.705679) (49, 0.704748) (50, 0.703829)
  };
  \addlegendentry{test2}
  \addplot[blue, mark=none,smooth, tension=0.6,line width=1pt, dashed] coordinates {
    (1, 3.10856) (2, 2.3508) (3, 1.7677) (4, 1.36502) (5, 1.11708) (6, 0.975115) (7, 0.894979) (8, 0.84818) (9, 0.819167) (10, 0.799961) (11, 0.786443) (12, 0.776395) (13, 0.768556) (14, 0.762179) (15, 0.756802) (16, 0.752133) (17, 0.747984) (18, 0.744229) (19, 0.740783) (20, 0.737586) (21, 0.734595) (22, 0.731778) (23, 0.729108) (24, 0.726567) (25, 0.724137) (26, 0.721804) (27, 0.719558) (28, 0.717391) (29, 0.715297) (30, 0.713273) (31, 0.71132) (32, 0.709443) (33, 0.707645) (34, 0.705931) (35, 0.704305) (36, 0.702766) (37, 0.701307) (38, 0.699922) (39, 0.698602) (40, 0.697339) (41, 0.696127) (42, 0.69496) (43, 0.693835) (44, 0.692748) (45, 0.691695) (46, 0.690675) (47, 0.689685) (48, 0.688723) (49, 0.687789) (50, 0.68688)
  };
  \addlegendentry{test3}

% Plot 2 (右上) - 无变化
\nextgroupplot
  \node[
    fill=gray!20,          % 浅灰色填充
    draw=none,              % 无边框
    align=left,             % 左对齐
    font=\small,            % 缩小字体
    anchor=north east,      % 锚点在右上角
    xshift=-3mm,            % 向左微调位置
    yshift=-1mm             % 向下微调位置
  ] at (rel axis cs:1,1) { % 定位在子图坐标系右上角
    Data Length: 2976\\     % 第一行
    Feature Length: 96       % 第二行
  };
  \addplot[red, mark=none, forget plot,smooth, tension=0.6,line width=1pt] coordinates {
    (1, 3.02829) (2, 2.32709) (3, 1.77932) (4, 1.38978) (5, 1.14045) (6, 0.992168) (7, 0.9061) (8, 0.855091) (9, 0.82334) (10, 0.802373) (11, 0.787704) (12, 0.776889) (13, 0.768535) (14, 0.761812) (15, 0.756205) (16, 0.751387) (17, 0.747143) (18, 0.743331) (19, 0.739854) (20, 0.736643) (21, 0.73365) (22, 0.730839) (23, 0.728181) (24, 0.725654) (25, 0.723241) (26, 0.720925) (27, 0.718694) (28, 0.716536) (29, 0.714442) (30, 0.712403) (31, 0.710414) (32, 0.708472) (33, 0.706578) (34, 0.704735) (35, 0.702951) (36, 0.701235) (37, 0.699593) (38, 0.69803) (39, 0.696545) (40, 0.695131) (41, 0.693781) (42, 0.692489) (43, 0.691246) (44, 0.690048) (45, 0.68889) (46, 0.68777) (47, 0.686684) (48, 0.685629) (49, 0.684603) (50, 0.683606)

  };
  \addplot[green, mark=none, forget plot,smooth, tension=0.6,line width=1.3pt,dotted] coordinates {
    (1, 3.19292) (2, 2.73044) (3, 2.33739) (4, 2.00701) (5, 1.73671) (6, 1.52028) (7, 1.35007) (8, 1.21777) (9, 1.1151) (10, 1.03495) (11, 0.97223) (12, 0.923295) (13, 0.885192) (14, 0.855389) (15, 0.832156) (16, 0.814135) (17, 0.799962) (18, 0.788578) (19, 0.779212) (20, 0.77128) (21, 0.76444) (22, 0.7584) (23, 0.752949) (24, 0.747925) (25, 0.743195) (26, 0.73868) (27, 0.734309) (28, 0.730046) (29, 0.725911) (30, 0.721969) (31, 0.718367) (32, 0.715141) (33, 0.712224) (34, 0.709574) (35, 0.707112) (36, 0.704808) (37, 0.702613) (38, 0.70053) (39, 0.698552) (40, 0.696665) (41, 0.694863) (42, 0.693158) (43, 0.691542) (44, 0.690011) (45, 0.688569) (46, 0.687198) (47, 0.685892) (48, 0.68465) (49, 0.68347) (50, 0.682352)

  };
  \addplot[blue, mark=none, forget plot,smooth, tension=0.6,line width=1pt, dashed] coordinates {
    (1, 3.02829) (2, 2.32709) (3, 1.77932) (4, 1.38978) (5, 1.14045) (6, 0.992168) (7, 0.9061) (8, 0.855091) (9, 0.82334) (10, 0.802373) (11, 0.787704) (12, 0.776889) (13, 0.768535) (14, 0.761812) (15, 0.756205) (16, 0.751387) (17, 0.747143) (18, 0.743331) (19, 0.739854) (20, 0.736643) (21, 0.73365) (22, 0.730839) (23, 0.728181) (24, 0.725654) (25, 0.723241) (26, 0.720925) (27, 0.718694) (28, 0.716536) (29, 0.714442) (30, 0.712403) (31, 0.710414) (32, 0.708472) (33, 0.706578) (34, 0.704735) (35, 0.702951) (36, 0.701235) (37, 0.699593) (38, 0.69803) (39, 0.696545) (40, 0.695131) (41, 0.693781) (42, 0.692489) (43, 0.691246) (44, 0.690048) (45, 0.68889) (46, 0.68777) (47, 0.686684) (48, 0.685629) (49, 0.684603) (50, 0.683606)

  };

% Plot 3 (左下) - 移除 xlabel 和 ylabel
\nextgroupplot
  \node[
    fill=gray!20,          % 浅灰色填充
    draw=none,              % 无边框
    align=left,             % 左对齐
    font=\small,            % 缩小字体
    anchor=north east,      % 锚点在右上角
    xshift=-3mm,            % 向左微调位置
    yshift=-1mm             % 向下微调位置
  ] at (rel axis cs:1,1) { % 定位在子图坐标系右上角
    Data Length: 12192\\     % 第一行
    Feature Length: 24       % 第二行
  };
  \addplot[red, mark=none, forget plot,smooth, tension=0.1,line width=1pt] coordinates {
   (1, 3.64764) (2, 2.02974) (3, 1.95808) (4, 1.9309) (5, 1.9123) (6, 1.89848) (7, 1.88753) (8, 1.87835) (9, 1.87061) (10, 1.86414) (11, 1.85866) (12, 1.85387) (13, 1.8496) (14, 1.84573) (15, 1.8422) (16, 1.83899) (17, 1.83607) (18, 1.83341) (19, 1.83097) (20, 1.82873) (21, 1.82666) (22, 1.82474) (23, 1.82295) (24, 1.82127) (25, 1.81969) (26, 1.8182) (27, 1.81679) (28, 1.81546) (29, 1.81419) (30, 1.81299) (31, 1.81183) (32, 1.81073) (33, 1.80967) (34, 1.80865) (35, 1.80767) (36, 1.80672) (37, 1.80581) (38, 1.80492) (39, 1.80406) (40, 1.80322) (41, 1.8024) (42, 1.80161) (43, 1.80082) (44, 1.80006) (45, 1.7993) (46, 1.79856) (47, 1.79782) (48, 1.7971) (49, 1.79637) (50, 1.79565)

  };
  \addplot[green, mark=none, forget plot,smooth, tension=0.6,line width=1.3pt,dotted] coordinates {
    (1, 4.81362) (2, 3.03236) (3, 2.3589) (4, 2.1119) (5, 2.02111) (6, 1.9796) (7, 1.95389) (8, 1.93523) (9, 1.92017) (10, 1.9062) (11, 1.89022) (12, 1.8753) (13, 1.86448) (14, 1.85586) (15, 1.84854) (16, 1.84271) (17, 1.8377) (18, 1.83346) (19, 1.82977) (20, 1.82655) (21, 1.82372) (22, 1.82124) (23, 1.81888) (24, 1.81634) (25, 1.81376) (26, 1.81105) (27, 1.80832) (28, 1.80584) (29, 1.80379) (30, 1.80271) (31, 1.80014) (32, 1.80084) (33, 1.79995) (34, 1.80447) (35, 1.8043) (36, 1.80523) (37, 1.80393) (38, 1.80254) (39, 1.8012) (40, 1.79988) (41, 1.79864) (42, 1.79745) (43, 1.79634) (44, 1.7953) (45, 1.79434) (46, 1.79344) (47, 1.79262) (48, 1.79189) (49, 1.79125) (50, 1.79068)

  };
  \addplot[blue, mark=none, forget plot,smooth, tension=0.1,line width=1pt, dashed] coordinates {
   (1, 3.64764) (2, 2.02974) (3, 1.95808) (4, 1.9309) (5, 1.9123) (6, 1.89848) (7, 1.88753) (8, 1.87835) (9, 1.87061) (10, 1.86414) (11, 1.85866) (12, 1.85387) (13, 1.8496) (14, 1.84573) (15, 1.8422) (16, 1.83899) (17, 1.83607) (18, 1.83341) (19, 1.83097) (20, 1.82873) (21, 1.82666) (22, 1.82474) (23, 1.82295) (24, 1.82127) (25, 1.81969) (26, 1.8182) (27, 1.81679) (28, 1.81546) (29, 1.81419) (30, 1.81299) (31, 1.81183) (32, 1.81073) (33, 1.80967) (34, 1.80865) (35, 1.80767) (36, 1.80672) (37, 1.80581) (38, 1.80492) (39, 1.80406) (40, 1.80322) (41, 1.8024) (42, 1.80161) (43, 1.80082) (44, 1.80006) (45, 1.7993) (46, 1.79856) (47, 1.79782) (48, 1.7971) (49, 1.79637) (50, 1.79565)

  };

% Plot 4 (右下) - 移除 xlabel
\nextgroupplot
  \node[
    fill=gray!20,          % 浅灰色填充
    draw=none,              % 无边框
    align=left,             % 左对齐
    font=\small,            % 缩小字体
    anchor=north east,      % 锚点在右上角
    xshift=-3mm,            % 向左微调位置
    yshift=-1mm             % 向下微调位置
  ] at (rel axis cs:1,1) { % 定位在子图坐标系右上角
    Data Length: 12192\\     % 第一行
    Feature Length: 96       % 第二行
  };
  \addplot[red, mark=none, forget plot,smooth, tension=0.1,line width=1pt] coordinates {
    (1, 3.66393) (2, 2.0372) (3, 1.96838) (4, 1.94069) (5, 1.92146) (6, 1.90726) (7, 1.8958) (8, 1.88618) (9, 1.87829) (10, 1.87172) (11, 1.86601) (12, 1.86096) (13, 1.85648) (14, 1.85251) (15, 1.84897) (16, 1.84579) (17, 1.8429) (18, 1.84026) (19, 1.83785) (20, 1.83561) (21, 1.83354) (22, 1.83161) (23, 1.82981) (24, 1.82811) (25, 1.82652) (26, 1.82501) (27, 1.82357) (28, 1.82221) (29, 1.8209) (30, 1.81965) (31, 1.81844) (32, 1.81728) (33, 1.81615) (34, 1.81504) (35, 1.81397) (36, 1.81292) (37, 1.81188) (38, 1.81086) (39, 1.80987) (40, 1.80889) (41, 1.80794) (42, 1.807) (43, 1.80609) (44, 1.8052) (45, 1.80432) (46, 1.80345) (47, 1.80259) (48, 1.80174) (49, 1.80089) (50, 1.80005)
  };
  \addplot[green, mark=none, forget plot,smooth, tension=0.6,line width=1.3pt,dotted] coordinates {
    (1, 4.82106) (2, 3.06852) (3, 2.43772) (4, 2.19935) (5, 2.0965) (6, 2.04418) (7, 2.01197) (8, 1.9884) (9, 1.96911) (10, 1.95067) (11, 1.93143) (12, 1.91551) (13, 1.90291) (14, 1.89215) (15, 1.88271) (16, 1.87366) (17, 1.86542) (18, 1.85805) (19, 1.85152) (20, 1.8458) (21, 1.84102) (22, 1.83714) (23, 1.83418) (24, 1.832) (25, 1.83038) (26, 1.82891) (27, 1.82727) (28, 1.82525) (29, 1.82243) (30, 1.81901) (31, 1.81559) (32, 1.81266) (33, 1.81048) (34, 1.80945) (35, 1.80922) (36, 1.80915) (37, 1.80846) (38, 1.80802) (39, 1.80845) (40, 1.80837) (41, 1.80698) (42, 1.80498) (43, 1.80364) (44, 1.80177) (45, 1.80066) (46, 1.78975) (47, 1.79102) (48, 1.80377) (49, 1.80611) (50, 1.78238)

  };
  \addplot[blue, mark=none, forget plot,smooth, tension=0.1,line width=1.3pt, dashed] coordinates {
    (1, 3.66393) (2, 2.0372)  (3, 1.96838) (4, 1.94069) (5, 1.92146) (6, 1.90726) (7, 1.8958) (8, 1.88618) (9, 1.87829) (10, 1.87172) (11, 1.86601) (12, 1.86096) (13, 1.85648) (14, 1.85251) (15, 1.84897) (16, 1.84579) (17, 1.8429) (18, 1.84026) (19, 1.83785) (20, 1.83561) (21, 1.83354) (22, 1.83161) (23, 1.82981) (24, 1.82811) (25, 1.82652) (26, 1.82501) (27, 1.82357) (28, 1.82221) (29, 1.8209) (30, 1.81965) (31, 1.81844) (32, 1.81728) (33, 1.81615) (34, 1.81505) (35, 1.81397) (36, 1.81292) (37, 1.81188) (38, 1.81086) (39, 1.80987) (40, 1.80889) (41, 1.80794) (42, 1.807) (43, 1.80609) (44, 1.8052) (45, 1.80432) (46, 1.80345) (47, 1.80259) (48, 1.80174) (49, 1.80089) (50, 1.80005)
  };
\end{groupplot}
  
    % 全局标签调整
    \node [below=0.4cm, font=\normalfont] at ($(group c1r2.south)!0.5!(group c2r2.south)$) {Epochs};
    \node [left=0.8cm, rotate=90, font=\normalfont] at ($(group c1r1.west)!0.5!(group c1r2.west)$) {Loss};
  
    % 紧凑型图例
    \node [draw=white,fill=white, 
           minimum width=3cm,        % 减小图例宽度
           minimum height=0.3cm,     % 减小图例高度
           above=0cm] at (current bounding box.north) {
      \begin{tikzpicture}
        \node[anchor=center] {
          \begin{tabular}{ccc}
            \tikz{\draw[red, thick] (0,0) -- (0.3cm,0);} Double &  % 缩短线段
            \hspace{0.3cm}\tikz{\draw[green, very thick, dotted] (0,0) -- (0.36cm,0);} Float &
            \hspace{0.3cm}\tikz{\draw[blue, thick, dashed] (0,0) -- (0.3cm,0);} Mixed \\
          \end{tabular}
        };
      \end{tikzpicture}
    };
  \end{tikzpicture}
  \vspace{-6pt}
\caption{The training-loss curves under different precision schemes, lengths of historical data, and lengths of feature input (PV ID = 2).}
\label{fig:training_curves}
\end{center}
\vspace{-12pt}
\end{figure}
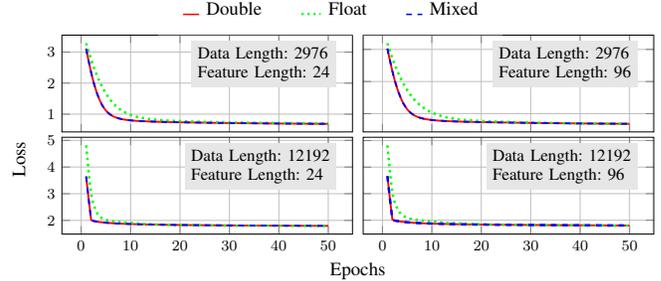

Compared to Table \ref{T5}, the time costs drop more than half when doing precision conversion via the ``Float” scheme and more than 1/3 when using the ``Mixed” scheme.  Fig. \ref{fig:training_curves} depicts the training loss curves under different settings. Note that the convergence patterns of the two reduced-precision schemes are similar to the default double-precision scheme.

% \begin{table}[htbp]
% \color{red}
% \centering
% \caption{Reduced-Precision On-Device Training of the LSTM Model (convert about 50\% variables and data)}
% \label{tab:comparison2}
% \resizebox{\columnwidth}{!}{%
% \begin{tabular}{c c c c}
% \toprule
% \makecell{\textbf{Total Data} \\ \textbf{Length}} & \makecell{\textbf{Feature} \\ \textbf{Length}} & \makecell{\textbf{On-Device} \\ \textbf{Time (sec)}} & \textbf{MAPE} \\
% \midrule
% 2976 & 24 & XX & YY\% \\
% 2976 & 96 & XX & YY\% \\
% 12192 & 24 & XX & YY\% \\
% 12192 & 96 & XX & YY\% \\
% \bottomrule
% \end{tabular}
% }
% \end{table}

\section{Conclusions}
In this paper, experiments of on-device model training are conducted in a smart meter for a PV power forecasting task. Results obtained from the on-device training approach are close to those on the PC in terms of the MAPE metric. Reduced-precision training techniques are devised for on-device training of the LSTM model, which can achieve about a 2X speed-up effect. Inspired by those encouraging results, training other machine learning models in the meter for more complicated applications will be the next step.

\end{document}